\documentclass{article}
\usepackage[final]{colm2026_conference}

\usepackage[T1]{fontenc}
\usepackage{microtype}
\usepackage{hyperref}
\usepackage{url}
\usepackage{booktabs}
\usepackage{tabularx}
\usepackage{amsmath}
\usepackage{amssymb}
\usepackage{graphicx}
\usepackage{algorithm}
\usepackage{algorithmic}
\usepackage{multirow}
\usepackage{xspace}
\usepackage{enumitem}
\usepackage{pifont}
\usepackage{subcaption}
\usepackage{wrapfig}
\usepackage{placeins}
\usepackage{tcolorbox}
\usepackage{listings}
\tcbuselibrary{skins, breakable}
\definecolor{takeawayframe}{HTML}{2A7F62}
\definecolor{takeawayback}{HTML}{EEF7F3}
\definecolor{takeawaytitle}{HTML}{1B5E44}
\newtcolorbox{takeaway}[1]{
  enhanced, breakable,
  colback=takeawayback, colframe=takeawayframe,
  colbacktitle=takeawaytitle, coltitle=white,
  fonttitle=\small\bfseries, fontupper=\small,
  title={#1},
  boxrule=0.8pt, arc=2pt,
  top=4pt, bottom=4pt, left=6pt, right=6pt,
  toptitle=3pt, bottomtitle=3pt
}

\usepackage{lineno}

\definecolor{darkblue}{rgb}{0, 0, 0.5}
\definecolor{softblue}{RGB}{100, 149, 237}
\hypersetup{colorlinks=true, citecolor=darkblue, linkcolor=softblue, anchorcolor=blue, urlcolor=darkblue}

\newif\ifrevisionmode
\ifdefined\showrevisions
  \revisionmodetrue
\else
  \revisionmodefalse
\fi
\ifrevisionmode
  \newcommand{\revision}[1]{\textcolor{red}{#1}}
  \newcommand{\revisioncolor}{\color{red}}
  \newcommand{\revisionlinks}{\hypersetup{citecolor=red,linkcolor=red,urlcolor=red}}
  \newcommand{\revisionurllinks}{\hypersetup{urlcolor=red}}
  \newcommand{\revisionlink}{\hypersetup{linkcolor=red}}
  \newcommand{\revisioncaptions}{\captionsetup{labelfont={color=red},textfont={color=red}}}
\else
  \newcommand{\revision}[1]{#1}
  \newcommand{\revisioncolor}{}
  \newcommand{\revisionlinks}{}
  \newcommand{\revisionurllinks}{}
  \newcommand{\revisionlink}{}
  \newcommand{\revisioncaptions}{}
\fi

\newcommand{\method}{\revision{\textit{CoEvoSkills}}\xspace}
\newcommand{\bench}{SkillsBench\xspace}
\newcommand{\skill}{\mathcal{S}}
\newcommand{\eg}{e.g.,\xspace}

\title{\revision{CoEvoSkills}: Self-Evolving Agent Skills via Co-Evolutionary Verification}

\author{%
\hspace{0pt}Hanrong Zhang$^{1}$\thanks{Equal contribution.} \quad
Shicheng Fan$^{1}$\footnotemark[1] \quad
Henry Peng Zou$^{1}$ \quad
Yankai Chen$^{2,3}$\thanks{Corresponding author: \texttt{yankaichen@acm.org}} \\
\hspace{0pt}\bf Zhenting Wang$^{2}$ \quad
Jiayu Zhou$^{4}$ \quad
Chengze Li$^{1}$ \quad
Wei-Chieh Huang$^{1}$ \quad
Yifei Yao$^{5}$ \\
\hspace{0pt}\bf Kening Zheng$^{1}$ \quad
Xue (Steve) Liu$^{2,3}$ \quad
Xiaoxiao Li$^{6}$ \quad
Philip S. Yu$^{1}$ \\[6pt]
\hspace{0pt}\mdseries
$^{1}$University of Illinois Chicago \quad
$^{2}$MBZUAI \quad
$^{3}$McGill University \\
$^{4}$Columbia University \quad
$^{5}$Zhejiang University \quad
$^{6}$University of British Columbia \\[4pt]
\texttt{\{hzhan135, psyu\}@uic.edu} \quad
\texttt{xiaoxiao.li@ece.ubc.ca}
}

\begin{document}

\ifcolmsubmission
\linenumbers
\fi

\maketitle

\begin{abstract}
Anthropic proposes the concept of skills for LLM agents to tackle multi-step professional tasks that simple tool invocations cannot address. A tool is a single, self-contained function, whereas a skill is a structured bundle of interdependent multi-file artifacts. Currently, skill generation is not only label-intensive due to manual authoring, but also may suffer from human--machine cognitive misalignment, which can lead to degraded agent performance, as evidenced by evaluations on SkillsBench. Therefore, we aim to enable agents to autonomously generate skills. However, existing self-evolving methods designed for tools cannot be directly applied to skills due to their increased complexity. To address these issues, we propose \revision{CoEvoSkills}, a self-evolving skills framework that enables agents to autonomously construct complex, multi-file skill packages. Specifically, \revision{CoEvoSkills} couples a Skill Generator that iteratively refines skills with a Surrogate Verifier that co-evolves to provide informative and actionable feedback without access to ground-truth test content. On SkillsBench, \revision{CoEvoSkills outperforms five baselines} on both Claude Code and Codex, and \revision{generalizes strongly} to six additional LLMs. \begingroup\revisionurllinks\revision{The code is publicly available at \url{https://github.com/Zhang-Henry/CoEvoSkills}.}\endgroup
\end{abstract}
\section{Introduction}
\label{sec:intro}

\begingroup\revisionlinks\revision{LLM agents have advanced rapidly in reasoning, planning, memory, and interaction with humans and their environments~\citep{yao2023react, zhang2025agent, luo2026agentauditor, zou2025llmbasedhumanagentcollaborationinteraction, huang2026rethinkingmemorymechanismsfoundation, wu2026gamhierarchicalgraphbasedagentic, zou2026userschangemindevaluating, huang2025deepresearchguard, min2025qucoragquantifyinguncertaintypretraining, min2025unihgkr, luo2026centaurevalbenchmarkinghumanintheloopvalue, li2026prefix, luo2026atelierevalagenticevaluationhumans, qiu2026yuvionvlmultimodalfoundation}.}\endgroup\ Among the key drivers of this progress is the ability to invoke external tools and APIs~\citep{schick2023toolformer, qin2024toolllm, patil2023gorilla}. However, professional open-ended tasks, such as complex software repair, multi-step scientific analysis, and enterprise data pipeline orchestration, require far more than isolated tool invocations. Agents must orchestrate a coherent procedure across multiple steps and artifacts: decomposing goals, coordinating tools, recovering from failures, and validating intermediate outputs. This is profoundly challenging because decisions are long-horizon, instructions and scripts are tightly coupled, and reliable environmental feedback is often sparse or delayed.

\begin{wrapfigure}{r}{0.7\linewidth}
\centering
\includegraphics[width=\linewidth]{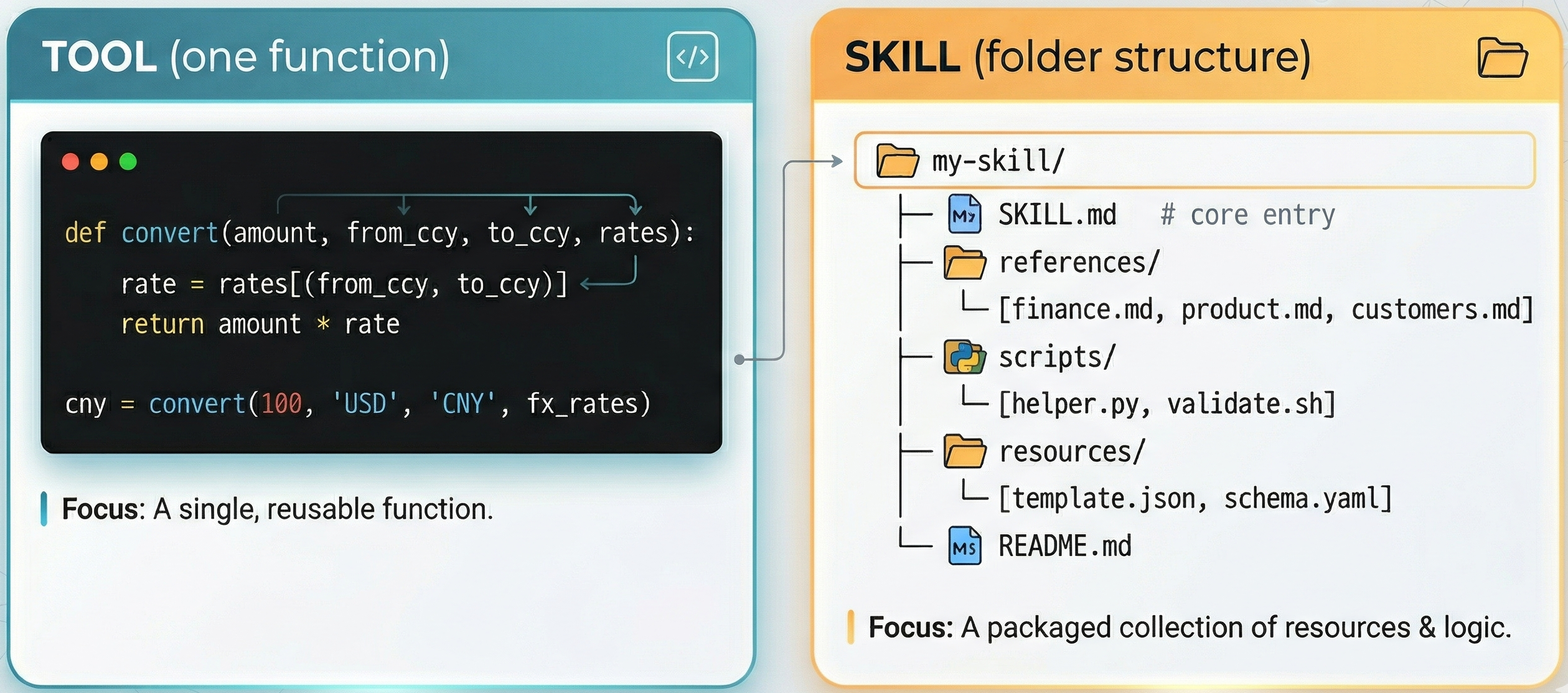}
\caption{Tool--skill difference illustration.}
\label{fig:tool_vs_skill_example}
\end{wrapfigure}

To bridge the gap, Anthropic proposed the concept of agent \emph{skills}~\citep{anthropic2025skills}. \autoref{fig:tool_vs_skill_example} illustrates the difference between a tool and a skill: a tool is usually a simple function, whereas a skill is a structured package of workflow instructions, executable scripts, and domain references~\citep{xu2026agentskills}. According to the systematic evaluation presented in \bench~\citep{li2026skillsbench}, equipping agents with well-crafted skills yields consistent performance gains across a broad spectrum of professional domains, including software engineering and scientific analysis etc., confirming that structured procedural guidance substantially augments task-solving capability beyond what bare tool access affords.

Despite the demonstrated utility of skills, the prevailing paradigm relies almost entirely on human authoring, a process that is \textbf{both labor-intensive and difficult to scale, with no systematic guarantee of output quality.} As demonstrated in \autoref{fig:domain_breakdown}, the \bench evaluation~\citep{li2026skillsbench} reveals that human-curated skills yield highly uneven gains: while certain domains benefit substantially, others, such as Natural Science, even exhibit degraded performance after skill integration. We hypothesize that a key driver of this inconsistency is \textbf{human--machine cognitive misalignment}: workflows and abstractions designed to be intuitive for human experts do not naturally match how LLM agents process context, reason, and act under execution constraints.

To reduce manual effort, recent approaches have shifted from pre-defining static tools or APIs to self-evolve tools or tools libraries by the LLM agent itself~\citep{chen2026skillcraftllmagentslearn,li2026yunjueagenttechreport,lu2026statictoolstesttimetool,wang2024voyager,xia2025live}. However, these methods suffer from a fundamental \emph{tool--skill gap}: they are inherently designed for one-shot generation of simple, self-contained functions, and \textbf{are inadequate for the creation of structured, multi-file skill packages} that coordinate workflow instructions, executable scripts, and domain references across multiple artifacts.
\revision{Moreover, EvoSkill provides ground-truth answers for failure diagnosis and uses held-out validation performance for skill selection~\citep{alzubi2026evoskillautomatedskilldiscovery}, limiting applicability in real-world settings where such supervision is unavailable.}

\begin{wrapfigure}{r}{0.65\linewidth}
\centering
\includegraphics[width=\linewidth]{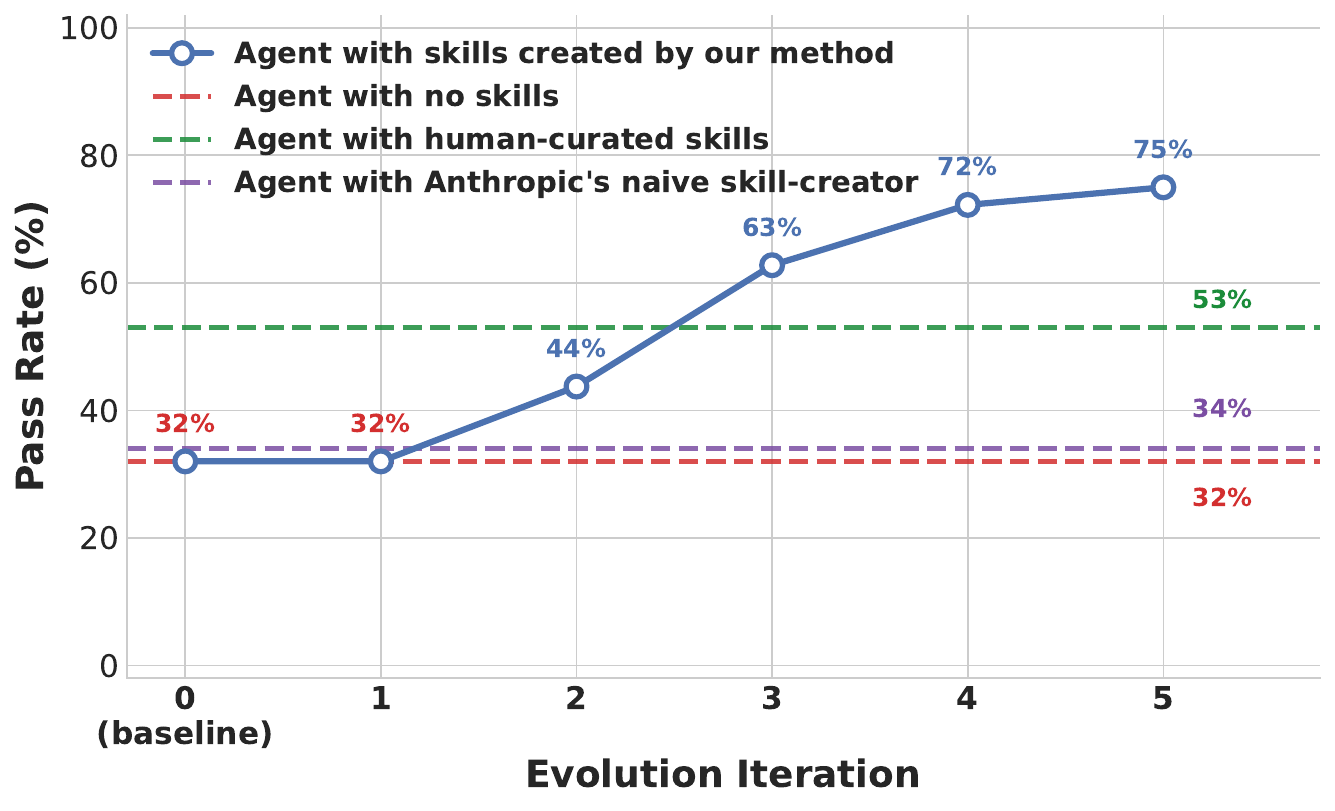}
\caption{Skill quality improvement across 5 evolution rounds. \method surpasses human-curated skills within 5 evolution iterations.}
\label{fig:evolution_trajectory}
\end{wrapfigure}
To address these challenges, we propose a self-evolving skills framework \method.
To overcome the inherent unreliability of one-shot multi-file skill generation, we design a main component, the \textbf{\texttt{Skill Generator}}, to iteratively generate and refine skill bundles, with skill quality steadily improving across evolution rounds, as shown in \autoref{fig:evolution_trajectory}. It also maintains a persistent conversation context that accumulates high-fidelity feedback from another main component, the \textbf{\texttt{Surrogate Verifier}}, across iterations. The \texttt{Surrogate Verifier}, a separate LLM session without inheriting the generator's biases, is designed to address the lack of ground-truth feedback in the real world. \revision{It synthesizes tests from task instructions, observable task inputs, and generated output artifacts, providing high-fidelity feedback that drives skill co-evolution.}

In summary, our key contributions are as follows:
\ding{182}~We introduce \method, a co-evolutionary framework for LLM agents to self-evolve robust multi-file skill packages.
\ding{183}~We demonstrate key insights on autonomous skill generation: (i)~\emph{agents create better skills than human-curated ones} by capturing the reasoning patterns and tool-use strategies that agents actually need; (ii)~\emph{self-evolved skills are portable across different model families}, as the evolved packages encode reusable task structure rather than model-specific artifacts.
\ding{184}~We conduct extensive experiments on \bench, \method achieves the highest pass rate of 71.1\% (+40.5pp over the no-skill baseline), substantially surpassing all five baselines. Furthermore, skills evolved by a single frontier LLM transfer effectively to six additional LLMs from five companies, yielding \revision{gains of approximately 35--44 percentage points} over their respective no-skill baselines.

\section{Related Work}
\label{sec:related}

\paragraph{LLM Agent Skills.}
Anthropic introduced \emph{Agent Skills}~\citep{anthropic2025skills} as shown in \autoref{fig:tool_vs_skill_example}.
A recent systematization~\citep{jiang2026sok} further distinguishes skills from atomic tools and one-off plans, defining them as reusable modules with explicit applicability and termination conditions.
\revision{\bench~\citep{li2026skillsbench} provides a systematic benchmark for evaluating reusable agent skills across diverse professional domains with deterministic task verifiers.}
Several learning-based approaches attempt to close this gap.
SAGE~\citep{wang2025sage} trains agents on chains of related tasks with a skill-integrated reward, yet the resulting skills remain single-file programmatic functions rather than structured multi-file packages.
SkillRL~\citep{xia2026skillrl} distills trajectories into a hierarchical skill bank via reinforcement learning, but relies on teacher-guided distillation and produces prompt-level heuristics rather than executable artifacts.
\method instead generates structured, multi-file skill packages and evolves them through iterative verification.

\paragraph{Self-evolving LLM Agents.}
A growing body of work automates the self-improvement of agent capabilities, yet existing methods exhibit two recurring shortcomings.
First, most self-evolving pipelines produce only single tools or function APIs or just prompt heuristics~\citep{wang2024voyager, li2026yunjueagenttechreport, xia2025live, lu2026statictoolstesttimetool, chen2026skillcraftllmagentslearn}, and none can construct the multi-file structure a full skill package demands.
Moreover, AutoSkill~\citep{yang2026autoskill} and AutoRefine~\citep{qiu2026autorefine} extract reusable knowledge as prompt templates rather than executable packages, while SEAgent~\citep{sun2025seagent} internalizes capabilities into model weights, making them non-inspectable and non-transferable~\citep{jiang2026sok}.
Second, some methods heavily rely on ground-truth signals for failure diagnosis, limiting applicability when such supervision is unavailable~\citep{alzubi2026evoskillautomatedskilldiscovery,sun2025seagent}.
\revision{\method addresses both limitations by evolving structured multi-file skill packages through isolated surrogate verification, yielding actionable diagnostics without ground-truth signals.}

\section{Method}
\label{sec:method}

\subsection{Method Overview}

\begin{figure}[t]
\centering
\includegraphics[width=\linewidth]{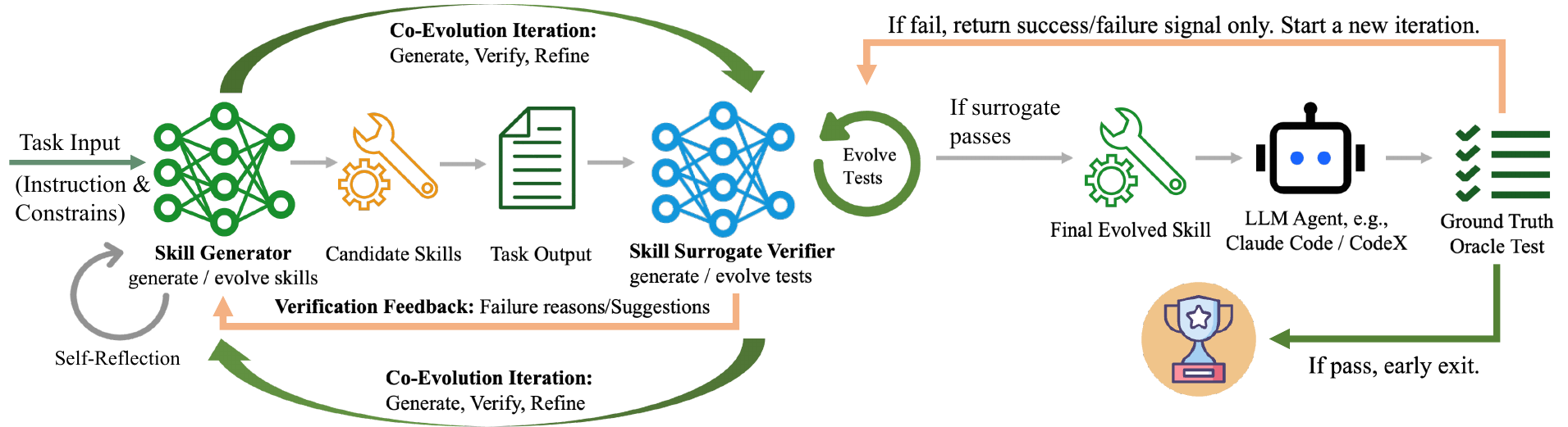}
\caption{Overview of the \method co-evolutionary framework. The \texttt{Skill Generator} and \texttt{Surrogate Verifier} co-evolve through iterative refinement. The verifier provides structured failure feedback to drive skill improvement, while a ground-truth oracle test returns only an opaque pass/fail signal, triggering test escalation and ensuring strict information isolation.}
\label{fig:evoSkill_framework}
\end{figure}

To answer the research question posed in \autoref{sec:intro}, two core difficulties must be overcome: (1)~generating a multi-file skill bundle in a single pass is inherently unreliable; and (2)~the agent lacks ground-truth feedback during self-evolution. \method addresses both challenges by co-evolving the agent's skill and a corresponding surrogate verifier. \autoref{fig:evoSkill_framework} illustrates the overall framework, and \autoref{alg:evolution} formalizes the complete co-evolutionary procedure. Given a task input, the \textbf{\texttt{Skill Generator}} produces candidate skills and executes them to obtain task outputs. An informationally isolated \textbf{\texttt{Surrogate Verifier}} then independently generates and evolves test assertions against those outputs, providing structured failure diagnostics back to the generator. The two components co-evolve through \textbf{iterative generate--verify--refine} cycles: whenever the surrogate tests pass, a ground-truth oracle test re-executes the skill in a fresh environment and returns only an opaque success/failure signal. If the oracle passes, the final evolved skill is deployed to the target LLM agent (\eg Claude Code~\citep{anthropic2025claudecode} and Codex~\citep{openai2025codex}); otherwise, the signal triggers a new co-evolution iteration, in which the verifier escalates its tests and the generator refines the skill accordingly. \revision{Next, we formalize \method in \autoref{sec:formulation} and \autoref{sec:evoskill}.}

\subsection{Problem Formulation}
\label{sec:formulation}

\paragraph{Task Definition.}
Since the LLM agent never knows what the held-out ground-truth tests actually check, i.e., the success criteria remain entirely hidden,
we define the task environment as a POMDP $\mathcal{M} = \langle \mathcal{X}, \mathcal{A}, \revision{\mathcal{T}}, \mathcal{O}, \Omega, \mathcal{R} \rangle$. Here, $\mathcal{X}$ is the underlying state space, i.e., the complete filesystem and processes, $\mathcal{A}$ comprises the agent's actions, i.e., terminal commands and file edits, $\revision{\mathcal{T}(x'\mid x,a)}$ is the deterministic state transition upon executing action $a$ in state $x$ and reaching successor state $x'$, $\mathcal{O}$ is the observation space, i.e., command execution results, $\Omega(o \mid x, a)$ maps post-action states to partial observations, and \revision{$y_H$ denotes the output artifacts contained in terminal state $x_H$, while $\mathcal{R}(y_H) \in [0,1]$ evaluates those artifacts against hidden ground-truth tests}. Because the agent only receives partial observations $o_t \sim \Omega(\cdot \mid x_t, a_t)$, it acts based on the observation--action history $h_t = (o_1, a_1, \ldots, a_{t-1}, o_t)$.

\paragraph{Optimization Objective.}
\label{sec:skill_structure}
Unlike atomic tools that offer simple functional interfaces, a \emph{skill} $\skill$~\citep{anthropic2025skills} is a structured bundle of domain-specific instructions, executable scripts, and reference materials that collectively guide an agent in navigating a task space. The skill conditions the agent's policy:
\begin{equation}
    a_t \sim \pi_\theta(a_t \mid h_t,\, \skill),
    \label{eq:policy}
\end{equation}
where $\pi_\theta$ is an LLM policy. We define the expected terminal reward under skill $\skill$ as:
\begin{equation}
    J(\skill) \;\triangleq\; \mathbb{E}_{\tau \sim P(\tau \mid \pi_\theta,\, \skill,\, \mathcal{M})} \bigl[\revision{\mathcal{R}(y_H)}\bigr],
    \label{eq:J}
\end{equation}
where \revision{$H$ is the terminal horizon and $\tau = (o_1, a_1, \ldots, a_{H-1}, o_H)$} is the execution trajectory. Our objective is to discover an optimal skill $\skill^{*}$ that maximizes $J$:
\begin{equation}
    \skill^{*} = \arg\max_{\skill}\; J(\skill).
    \label{eq:objective}
\end{equation}

\begin{algorithm}[t]
\caption{\method co-evolution Algorithm}
\label{alg:evolution}
\small
\begin{algorithmic}[1]
\REQUIRE Instruction $I$, \revision{observable task inputs $\mathcal{D}$, background knowledge $\mathcal{B}$,} environment $\mathcal{E}$, meta-skill $\skill_{\mathrm{meta}}$ (\texttt{skill-creator})
\REQUIRE \revision{Ground-truth oracle rounds $K{=}5$, surrogate repair retries $M{=}15$}, context cap $\beta{=}0.7$
\REQUIRE LLM policy $\pi_\theta$ (generator), independent verifier policy $\pi_\theta^{V}$
\ENSURE Final evolved skill $\skill^*$
\STATE $C \leftarrow (I,\, \revision{\mathcal{B},\,} \skill_{\mathrm{meta}})$ \hfill \texttt{//} in: instruction $I$, \revision{background knowledge $\mathcal{B}$,} meta-skill; out: initial context $C$
\STATE $\skill^{(0)} \sim \pi_\theta\!\left(\cdot \mid C\right)$ \hfill \texttt{//} in: context $C$; out: skill bundle $\skill^{(0)}$ (code + SKILL.md)
\STATE $\mathcal{V}^{(0)} \leftarrow \revision{\bot}$ \hfill \texttt{//} \revision{uninitialized surrogate verifier test suite}
\STATE $i \leftarrow 0$;\; $j \leftarrow 0$;\; $n \leftarrow 0$;\; $r \leftarrow 0$ \hfill \texttt{//} \revision{skill ver.; test ver.; oracle rounds; repair retries}
\STATE $\mathcal{R}_{\mathrm{best}} \leftarrow 0$;\; $\skill^* \leftarrow \skill^{(0)}$ \hfill \texttt{//} $\mathcal{R}_{\mathrm{best}}$: best oracle score so far
\WHILE{$n < \revision{K}$ \textbf{and} $r < M$}
    \STATE \texttt{//} \textit{--- \texttt{Skill Generator}: execute and produce outputs ---}
    \STATE $\revision{y^{(i)}} \leftarrow \Phi(\skill^{(i)},\, \mathcal{E})$ \hfill \texttt{//} $\Phi$: execute skill in env and collect outputs; out: artifacts $\revision{y^{(i)}}$
    \IF{LLM context usage proportion $> \beta$}
        \STATE \textbf{break} \hfill \texttt{//} prevent LLM context overflow
    \ENDIF
    \IF{\revision{$j=0$ \textbf{and} $\mathcal{V}^{(0)}=\bot$}}
        \STATE \revision{$\mathcal{V}^{(0)} \sim \pi_\theta^{V}\!\left(\cdot \mid I,\, \mathcal{D},\, y^{(i)}\right)$} \hfill \revision{\texttt{//} initialize tests from instruction, observable inputs, and artifacts}
    \ENDIF
    \STATE \texttt{//} \textit{--- \texttt{Surrogate Verifier}: evaluate and refine (\autoref{eq:proxy_reward}, \autoref{eq:skill_opt})\ ---}
    \STATE $\tilde{\mathcal{R}}^{(i,j)} \leftarrow \tilde{\mathcal{R}}(\revision{y^{(i)}},\, \mathcal{V}^{(j)})$ \hfill \texttt{//} $\tilde{\mathcal{R}}$: surrogate reward; in: artifacts, tests; out: pass rate $\in [0,1]$
    \IF{$\tilde{\mathcal{R}}^{(i,j)} < 1$}
        \STATE $\mathcal{F}^{(i,j)} \sim \pi_\theta^{V}\!\left(\cdot \mid I,\, \revision{\mathcal{D},\, y^{(i)}},\, \mathcal{V}^{(j)}\right)$ \hfill \texttt{//} in: $I$, \revision{inputs $\mathcal{D}$,} artifacts $\revision{y^{(i)}}$, tests $\mathcal{V}^{(j)}$; out: diagnostic $\mathcal{F}$
        \STATE $C \leftarrow C \oplus \mathcal{F}^{(i,j)}$ \hfill \texttt{//} append error diagnostic $\mathcal{F}$ to generator context $C$
        \STATE $\skill^{(i+1)} \sim \pi_\theta\!\left(\cdot \mid \skill^{(i)},\; C\right)$ \hfill \texttt{//} skill refinement (\autoref{eq:skill_update})
        \STATE $i \leftarrow i{+}1$;\; $r \leftarrow r{+}1$;\; \textbf{continue} \hfill \texttt{//} evolve $\skill$; $\mathcal{V}^{(j)}$ locked
    \ENDIF
    \STATE \texttt{//} \textit{--- Ground-Truth Oracle Test: independent re-execution in fresh environment ---}
    \STATE $\revision{\hat{y}^{(i)}} \leftarrow \Phi(\skill^{(i)},\, \mathcal{E}')$ \hfill \texttt{//} in: skill $\skill^{(i)}$, fresh env $\mathcal{E}'$; out: artifacts $\revision{\hat{y}^{(i)}}$
    \STATE $\mathcal{R}^{(i)} \leftarrow \mathcal{R}(\revision{\hat{y}^{(i)}})$;\; $n \leftarrow n{+}1$ \hfill \texttt{//} $\mathcal{R}$: ground-truth oracle reward; out: score $\in [0,1]$
    \IF{$\mathcal{R}^{(i)} = 1$}
        \STATE $\skill^* \leftarrow \skill^{(i)}$;\; \textbf{return} $\skill^*$ \hfill \texttt{//} early exit: perfect score
    \ELSIF{$\mathcal{R}^{(i)} > \mathcal{R}_{\mathrm{best}}$}
        \STATE $\mathcal{R}_{\mathrm{best}} \leftarrow \mathcal{R}^{(i)}$;\; $\skill^* \leftarrow \skill^{(i)}$ \hfill \texttt{//} save best snapshot
    \ENDIF
    \STATE \texttt{//} \textit{--- Co-evolution: test escalation (\autoref{eq:test_opt}, \autoref{eq:verifier_update}) ---}
    \STATE $C \leftarrow C \oplus \mathbf{1}[\mathcal{R}^{(i)} {<} 1]$ \hfill \texttt{//} $\mathbf{1}[\cdot]$: indicator fn; append oracle pass/fail bit to $C$ (no test content)
    \STATE $\mathcal{V}^{(j+1)} \sim \pi_\theta^{V}\!\left(\cdot \mid I,\, \revision{\mathcal{D},\, y^{(i)}},\, \mathcal{V}^{(j)}\right)$ \hfill \texttt{//} verifier escalation (\autoref{eq:verifier_update})
    \STATE $j \leftarrow j{+}1$ \hfill \texttt{//} $\mathcal{V}$ evolves; $\skill^{(i)}$ re-evaluated next iteration
\ENDWHILE
\STATE \textbf{return} $\skill^*$ \hfill \texttt{//} save best skill
\end{algorithmic}
\end{algorithm}

\subsection{\method Framework}
\label{sec:evoskill}

However, directly optimizing $J(\skill)$ (\autoref{eq:objective}) is intractable: ground-truth evaluation is computationally expensive and returns only an opaque pass/fail signal; no test content or failure details are revealed to the agent. To provide dense, actionable feedback, we introduce a surrogate verifier reward $\tilde{\mathcal{R}}(\revision{y}, \mathcal{V})$, defined by a suite of deterministic test assertions $\mathcal{V} = \{e_1, \ldots, e_{|\mathcal{V}|}\}$ generated by an independent verifier
\begin{equation}
    \tilde{\mathcal{R}}(\revision{y}, \mathcal{V}) \;\triangleq\; \frac{1}{|\mathcal{V}|}\sum_{k=1}^{|\mathcal{V}|}\mathbf{1}\!\left[e_k(\revision{y})\right] \;\in\; [0,\, 1],
    \label{eq:proxy_reward}
\end{equation}
where \revision{$y$ denotes the output artifacts produced by skill execution and $\mathbf{1}[e_k(y)]$ indicates whether assertion $e_k$ passes on them}. However, the surrogate is only useful insofar as it faithfully approximates the hidden $\mathcal{R}$. This creates a coupled optimization problem: the skill must maximize a proxy that itself must be aligned with the hidden ground truth. Since the ground-truth (GT) oracle test reveals only a binary pass/fail signal $\mathbf{1}[\mathcal{R}(\revision{\hat{y}^{(i)}}) < 1]$, where $\revision{\hat{y}^{(i)}}$ denotes the oracle's independent re-execution output, and no test content, we cannot directly optimize against $\mathcal{R}$. Instead, letting $I$ denote the task instruction, we define the rollout operator $\Phi(\skill, \mathcal{E})$ as the execution output obtained by rolling out $\pi_\theta(\cdot \mid h_t, \skill)$ in environment $\mathcal{E}$, so that $\revision{y^{(i)}} = \Phi(\skill^{(i)}, \mathcal{E})$ is the output of the $i$-th skill version and $\revision{\hat{y}^{(i)}} = \Phi(\skill^{(i)}, \mathcal{E}')$ is the output produced by an independent oracle re-execution in a fresh environment $\mathcal{E}'$. $\mathcal{V}^{(j)}$ is the $j$-th version of the surrogate verifier test suite. \revision{We use $\mathcal{D}$ for all inputs available during skill evolution, including instance data and background documents $\mathcal{B}$ ($\mathcal{B} \subseteq \mathcal{D}$); $\mathcal{D}$ contains no hidden oracle information.} \method~proceeds via alternating refinement:
\begin{align}
    \textit{Skill refinement:} \quad & \skill^{(i+1)} \;\leftarrow\; \arg\max_{\skill}\; \tilde{\mathcal{R}}\!\bigl(\Phi(\skill,\,\mathcal{E}),\; \mathcal{V}^{(j)}\bigr), \label{eq:skill_opt} \\
    \textit{Test escalation:} \quad & \mathcal{V}^{(j+1)} \sim \pi_\theta^{V}\!\bigl(\,\cdot \mid I,\; \revision{\mathcal{D},\; y^{(i)}},\; \mathcal{V}^{(j)}\bigr), \;\; \text{if } \mathbf{1}\!\left[\tilde{\mathcal{R}}(\revision{y^{(i)}},\, \mathcal{V}^{(j)}) {=} 1 \;\wedge\; \mathcal{R}(\revision{\hat{y}^{(i)}}) {<} 1\right] \label{eq:test_opt}
\end{align}
In practice, the $\arg\max$ in \autoref{eq:skill_opt} is approximated by iterative LLM sampling (\autoref{eq:skill_update}). Skill refinement maximizes $\tilde{\mathcal{R}}$ under a fixed verifier test suite $\mathcal{V}^{(j)}$; test escalation is triggered only when the oracle's binary signal exposes a gap between $\tilde{\mathcal{R}}$ and $\mathcal{R}$, forcing the verifier to independently strengthen its tests without any access to ground-truth test content. \method realizes this alternating optimization through three informationally isolated components: a \emph{\texttt{Skill Generator}} and a \emph{\texttt{Surrogate Verifier}} orchestrated in a co-evolutionary loop (\autoref{alg:evolution}).

\paragraph{\texttt{Skill Generator}.}
A single-pass skill generation often produces bundles with coverage gaps and logical errors, because the agent lacks ground-truth feedback during generation. To enable iterative refinement, the \texttt{Skill Generator} maintains a persistent conversation context $C$. \revision{We study \emph{agent-driven skill evolution}, in which the generator begins with task-relevant background knowledge available at initialization, such as organizational documentation or domain references. Accordingly, $C^{(0)} = (I, \mathcal{B}, \skill_{\mathrm{meta}})$, where $I$ is the task instruction, $\mathcal{B}$ denotes this task-relevant background knowledge, and $\skill_{\mathrm{meta}}$ is a domain-agnostic meta-skill (\texttt{skill-creator}) that teaches how to create skills. At each revision, the LLM $\pi_\theta$ (\autoref{eq:policy}) updates the current skill $\skill^{(i)}$ using accumulated surrogate-verifier diagnostics:}
\begin{equation}
    \skill^{(i+1)} \sim \pi_\theta\!\left(\cdot \mid \skill^{(i)},\; C^{(i+1)}\right), \quad
    C^{(i+1)} = C^{(i)} \oplus \mathcal{F}^{(i,j)},
    \label{eq:skill_update}
\end{equation}
where $\skill^{(i)}$ is the $i$-th skill version, $\mathcal{F}^{(i,j)}$ is the failure diagnostic from the \texttt{Surrogate Verifier} after evaluating $\skill^{(i)}$ under test suite $\mathcal{V}^{(j)}$, comprising failed test cases, root-cause analysis, and actionable revision suggestions, and $\oplus$ appends detailed feedback to the LLM context. The agent executes $\skill^{(i)}$ via rollout $\revision{y^{(i)}} = \Phi(\skill^{(i)}, \mathcal{E})$ (line~8), and the resulting outputs are passed to the \texttt{Surrogate Verifier} for evaluation.

\paragraph{\texttt{Surrogate Verifier}.}
Since the ground-truth reward $\mathcal{R}$ returns only an opaque pass/fail signal, the \texttt{Skill Generator} lacks dense feedback to diagnose and correct errors. The \texttt{Surrogate Verifier} addresses this gap by serving as a proxy for $\mathcal{R}$, providing per-assertion failure diagnostics that the opaque oracle cannot. \revision{To reduce confirmation bias, the \texttt{Surrogate Verifier} uses an independent LLM session $\pi_\theta^V$ and receives only $I$, $\mathcal{D}$ (including $\mathcal{B}$), $y^{(i)}$, and $\mathcal{V}^{(j)}$. It has no access to generator reasoning or skill source, or to hidden oracle criteria, test content, or diagnostics, while remaining able to construct input-grounded tests.}
The verifier generates a proxy verifier test suite $\mathcal{V} = \{e_1, \ldots, e_{|\mathcal{V}|}\}$ of deterministic assertions, yielding the proxy reward $\tilde{\mathcal{R}}(\revision{y}, \mathcal{V})$ defined in \autoref{eq:proxy_reward}. \revision{After the first rollout, it initializes $\mathcal{V}^{(0)}$ from $(I, \mathcal{D}, y^{(0)})$ before computing the first surrogate reward (\autoref{alg:evolution}, lines~12--16).} It then iteratively refines this verifier test suite by reading the previous script $\mathcal{V}^{(j)}$ and current outputs $\revision{y^{(i)}}$:
\begin{equation}
    \mathcal{V}^{(j+1)} \sim \pi_\theta^{V}\!\bigl(\,\cdot \mid I,\; \revision{\mathcal{D},\; y^{(i)}},\; \mathcal{V}^{(j)}\bigr),
    \label{eq:verifier_update}
\end{equation}
where the verifier conditions on the task instruction $I$, \revision{observable task inputs $\mathcal{D}$,} the agent's current outputs $\revision{y^{(i)}}$, and its own previous test script $\mathcal{V}^{(j)}$ to produce an improved verifier test suite $\mathcal{V}^{(j+1)}$. When the surrogate reward indicates failure ($\tilde{\mathcal{R}} < 1$), the verifier additionally generates a structured failure diagnostic $\mathcal{F}^{(i,j)} \sim \pi_\theta^{V}(\cdot \mid I,\, \revision{\mathcal{D},\, y^{(i)}},\, \mathcal{V}^{(j)})$, including per-assertion results, root-cause analysis, and actionable revision suggestions fed back to the \texttt{Skill Generator} (\autoref{eq:skill_update}).
\paragraph{Co-evolution of \texttt{Skill Generator} and \texttt{Surrogate Verifier}.}
The alternating optimization in \autoref{eq:skill_opt}--\autoref{eq:test_opt} couples two feedback pathways (\autoref{alg:evolution}). When the surrogate reward indicates failure ($\tilde{\mathcal{R}} < 1$), the verifier test suite $\mathcal{V}$ is held fixed and the failure diagnostic $\mathcal{F}$ drives skill revision (\autoref{eq:skill_update}). When the surrogate test passes but the ground-truth oracle fails, only an opaque pass/fail bit is returned, i.e., no test content or failure details to prevent the \texttt{Skill Generator} from overfitting to the held-out tests. The \texttt{Surrogate Verifier} must then independently escalate its test suite (\autoref{eq:verifier_update}) according to \revision{the observable task inputs and current output artifacts}. For example, it may generate more diverse, comprehensive and challenging test cases. Through this dual-feedback mechanism, the skill $\skill$ improves under surrogate test pressure. \revision{\autoref{fig:evolution_trajectory} confirms convergence within few co-evolutionary iterations.}

\section{Experiments}
\label{sec:experiments}
\revision{In this section, we answer three Research Questions (RQs): (1) Are self-evolved skills useful, and can they outperform human-curated skills (\autoref{sec:main_results})? (2) Can evolved skills transfer across models and agent harnesses (\autoref{sec:cross_model})? (3) How are performance gains distributed across professional domains (\autoref{sec:domain})? Component ablations are reported in \autoref{app:ablation}.}

\subsection{Experimental Setup}
\label{sec:bench}
We evaluate \method on \bench~\citep{li2026skillsbench}. \revision{Our experiments cover 85 tasks,} providing broad coverage of the real-world distribution of skill-augmented tasks. Each task is paired with a deterministic verifier, enabling reproducible, binary pass/fail evaluation without subjective human judgment.
We adopt \bench as the sole evaluation suite because to the best of our knowledge, it is the only benchmark purpose-built for assessing the utility of agent \emph{skills}.
The primary metric is \textbf{pass rate}: the proportion of tasks with reward $=1.0$, i.e., all tests passed, otherwise reward $=0.0$. Unless otherwise stated, all conditions use the same instruction format. For conditions with pre-installed skills, the task instruction notes their availability.

\begin{wrapfigure}{r}{0.55\linewidth}
\centering
\includegraphics[width=\linewidth]{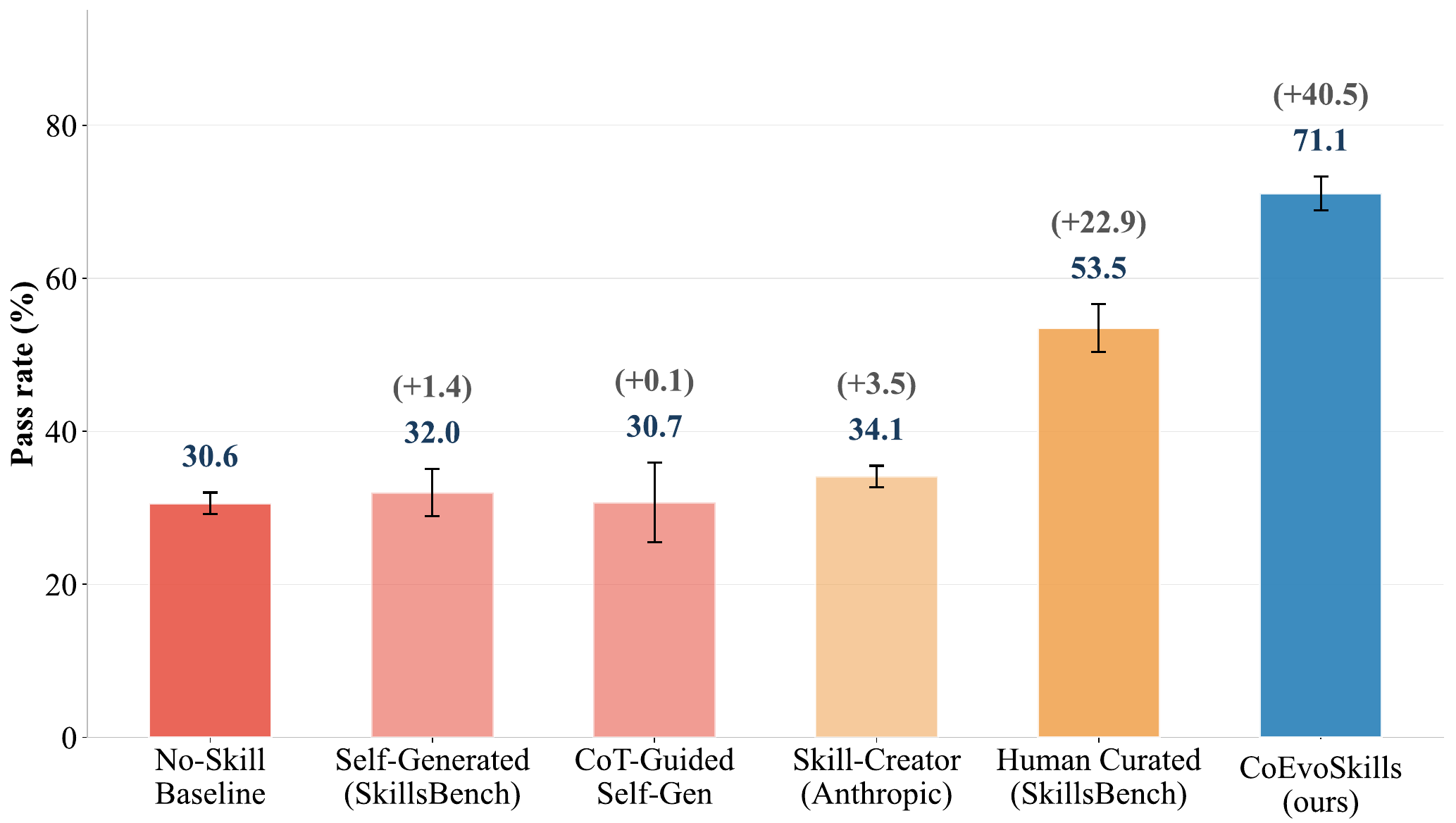}
\caption{Skill quality comparisons with baselines on \bench (Claude Opus 4.6 + Claude-Code). Error bars: $\pm$1 std over 5 runs.}
\label{fig:opus_main}
\end{wrapfigure}

We compare \revision{five} baselines against \method. The \textbf{No-Skill Baseline} evaluates the agent's performance when no skills are available. \textbf{Self-Generated Skills} replicates the one-pass self-generation condition from \bench~\citep{li2026skillsbench}: the agent generates one to five skill documents in a single pass before solving the task, with no iterative evolution or verification (see prompt in \autoref{app:prompt_selfgen}). \textbf{CoT-Guided Self-Generation} extends the Self-Generated Skills condition with a structured five-step chain-of-thought prompt (see prompt in \autoref{app:prompt_cot}). \textbf{Skill-Creator} adopts Anthropic's official \texttt{skill-creator}~\citep{anthropic2025skills}. Since our evaluation is fully autonomous, we replace its human-interactive steps with autonomous equivalents (see \autoref{app:prompt_sc}): a first session iteratively drafts, self-tests, and refines skills for at least three iterations, and a second session solves the task using the resulting skills. \textbf{Human Curated Skills} pre-installs the human-authored skill packages released with \bench. For the evolution agent, we use Claude Opus 4.6 and GPT-5.2 as backbone models. \revision{Each evolution run starts with general background knowledge $\mathcal{B}$ available at initialization, without task-instance solution procedures or hidden oracle information.}
For baseline comparisons, each primary method is evaluated over 5 independent runs, and mean $\pm$ standard deviation is reported.

In addition, we also evaluate the transferability of the skills evolved by Claude Opus 4.6 on six additional models: GPT-5.2~\citep{openai2025gpt5}, Claude Sonnet 4.5~\citep{anthropic2025sonnet45}, Claude Haiku 4.5~\citep{anthropic2025haiku45}, Qwen3-Coder-480B~\citep{qwen2025qwen3}, DeepSeek V3-671B~\citep{deepseek2024v3} and Mistral Large 3-675B~\citep{mistral2025large3}. Each model is evaluated over 3 independent runs.

\begingroup
\revisioncolor
\revisionlinks
\begin{table}[ht]
\revisioncolor
\revisioncaptions
\centering
\small
\begin{tabular}{@{}p{0.3\linewidth}p{0.65\linewidth}@{}}
\toprule
\textbf{Component} & \textbf{Setting} \\
\midrule
Backbones & Claude Opus 4.6, GPT-5.2 \\
Surrogate verifier model & Same as evolution backbone (Claude Opus 4.6 / GPT-5.2) \\
Ground-truth oracle test agent & \texttt{Claude-Code} (Claude Opus 4.6) / \texttt{Codex} (GPT-5.2) \\
Evolution stage & \revision{$K{=}5$ Ground Truth Oracle rounds, $M{=}15$ surrogate repair retries} \\
Evolution runtime & timeout multiplier $5\times$ (effective 3000s/task), 4 parallel workers \\
Evaluation stage & timeout 7200s/task, 10 parallel workers \\
\bottomrule
\end{tabular}
\caption{Shared configuration for evolution and evaluation.}
\label{tab:setup}
\end{table}

\begin{table}[ht]
\revisioncolor
\revisioncaptions
\centering
\small
\begin{tabular}{@{}p{0.3\linewidth}p{0.4\linewidth}@{}}
\toprule
\textbf{Model} & \textbf{Agent harness} \\
\midrule
Claude Opus 4.6 & \texttt{Claude-Code}~\citep{anthropic2025claudecode} \\
GPT-5.2 & \texttt{Codex}~\citep{openai2025codex} \\
Claude Sonnet 4.5 & \texttt{Claude-Code}~\citep{anthropic2025claudecode} \\
Claude Haiku 4.5 & \texttt{Terminus-2}~\citep{merrill2026terminalbench} \\
Qwen3 Coder & \texttt{Terminus-2}~\citep{merrill2026terminalbench} \\
DeepSeek V3 & \texttt{Terminus-2}~\citep{merrill2026terminalbench} \\
Mistral Large 3 & \texttt{Terminus-2}~\citep{merrill2026terminalbench} \\
\bottomrule
\end{tabular}
\caption{Agent harness used for each model in ground-truth oracle evaluation.}
\label{tab:agent_harness}
\end{table}
\endgroup

\FloatBarrier

\subsection{RQ1: Skill Quality Comparison}
\label{sec:main_results}

\autoref{fig:opus_main} presents the core comparison on Claude Opus 4.6 with Claude-Code. \method reaches a \textbf{71.1\%} pass rate, exceeding the no-skill baseline (30.6\%) by $+40.5$pp and human-curated skills (53.5\%) by $+17.6$pp. The Skill-Creator baseline achieves only 34.1\%, barely above the no-skill baseline. Two in-session self-generation baselines perform no better: the \bench self-generated skills baseline~\citep{li2026skillsbench} reaches 32.0\% ($\pm$3.1), and a CoT-guided variant that follows a structured five-step chain-of-thought prompt reaches 30.7\% ($\pm$5.2), both providing negligible improvement over the no-skill baseline. This confirms that skill generation without co-evolutionary verification is insufficient regardless of the generation strategy: the gains of \method originate from the iterative verification loop, not from the skill-creation prompt itself. We summarize the main insight as follows.

\begin{takeaway}{Takeaway 1: Agents can create better skills than humans}
Self-evolved skills outperform human-curated ones by encoding agent-native reasoning patterns, tool-use preferences, and decomposition strategies rather than following human assumptions about how agents should operate.
\end{takeaway}
\vspace{-1em}

\subsection{RQ2: Skill Cross-Model Transferability}
\label{sec:cross_model}

\begin{figure}[t]
\centering
\includegraphics[width=\linewidth]{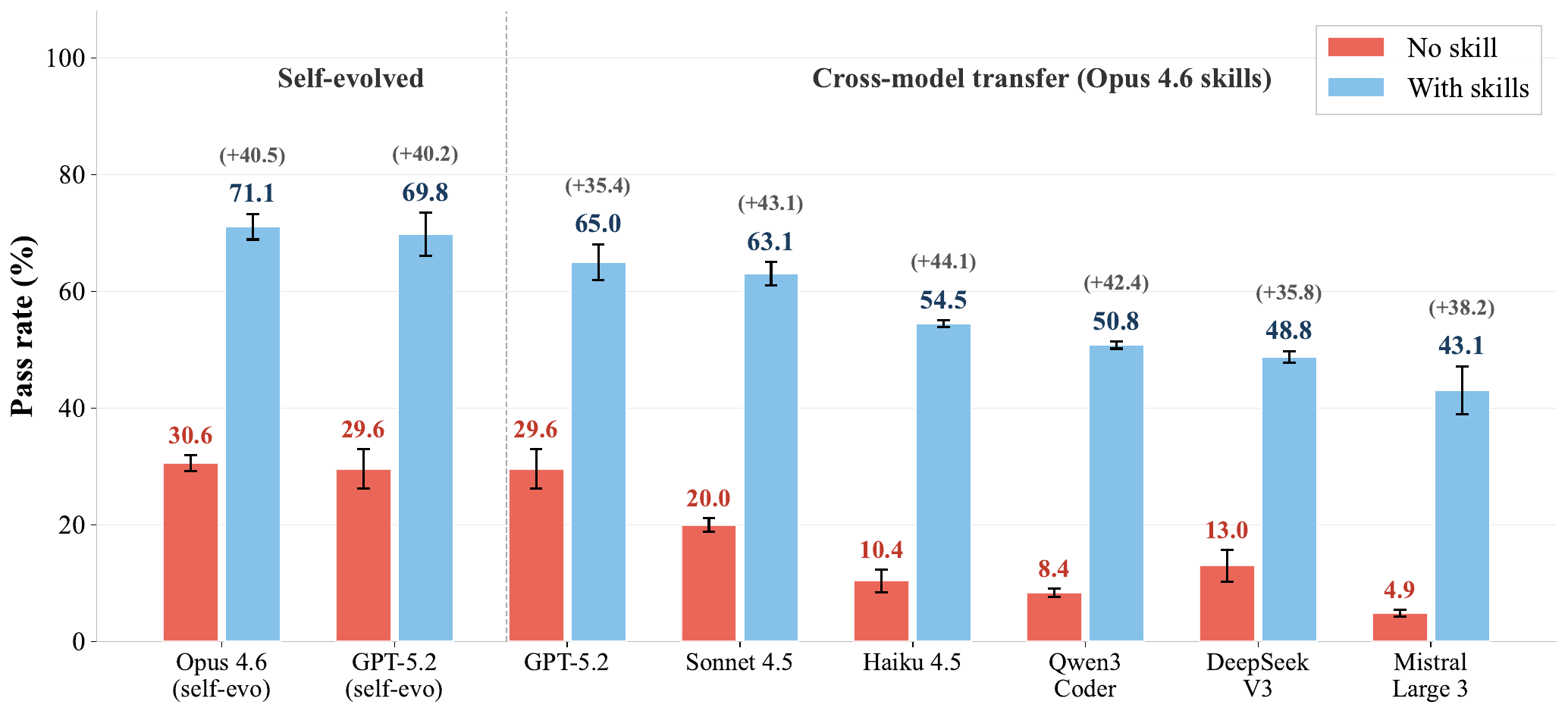}
\caption{Cross-model skill transferability on \bench. Skills evolved by Claude Opus 4.6 are transferred to six additional models spanning five providers. Each pair of bars shows the no-skill baseline (red) and the with-skills pass rate (blue). Delta annotations indicate absolute improvement. All models benefit substantially, with \revision{gains of approximately 35--44 percentage points across models}, confirming that the evolved skills encode reusable task structure rather than model-specific artifacts.}
\label{fig:cross_model}
\end{figure}
Having established that \method produces high-quality skills on two primary LLMs, we next examine whether these skills generalize across LLM model families. As shown in \autoref{fig:cross_model}, self-evolved skills yield substantial gains on both primary backbones: Claude Opus 4.6 (+40.5pp) and GPT-5.2 (+40.2pp). Transferring Opus-evolved skills to six additional models yields \revision{gains of approximately 35--44 percentage points across models}, demonstrating that the distilled workflows are not tied to the originating model. \begingroup\revisionlink\revision{Exact pass rates are reported in \autoref{tab:cross_model}.}\endgroup

Interestingly, GPT-5.2 benefits from Opus-transferred skills (65.0\%), yet its own self-evolved skills (69.8\%) still outperform the transferred set by 4.8pp, indicating a modest but consistent advantage for model-matched evolution. We summarize the main takeaway as follows.

\begingroup
\revisioncolor
\begin{table}[t]
\revisioncolor
\revisioncaptions
\centering
\small
\setlength{\tabcolsep}{5pt}
\begin{tabular}{@{}lccc@{}}
\toprule
\textbf{Model} & \textbf{With skills} & \textbf{No skill} & \textbf{$\Delta$} \\
\midrule
\multicolumn{4}{@{}l}{\textbf{Self-Evolved Skills}} \\
Claude Opus 4.6 (self-evolved) & 71.1 & 30.6 & +40.5 \\
GPT-5.2 (self-evolved) & 69.8 & 29.6 & +40.2 \\
\midrule
\multicolumn{4}{@{}l}{\textbf{Cross-Model Transfer (Opus 4.6 Evolved Skills)}} \\
GPT-5.2 & 65.0 & 29.6 & +35.4 \\
Claude Sonnet 4.5 & 63.1 & 20.0 & +43.1 \\
Claude Haiku 4.5 & 54.5 & 10.4 & +44.1 \\
Qwen3 Coder & 50.8 & 8.4 & +42.4 \\
DeepSeek V3 & 48.8 & 13.0 & +35.8 \\
Mistral Large 3 & 43.1 & 4.9 & +38.2 \\
\bottomrule
\end{tabular}
\caption{Cross-model skill transferability, pass rate (\%).}
\label{tab:cross_model}
\end{table}
\endgroup

\begin{takeaway}{Takeaway 2: Skills are portable across model families}
Self-evolved skills encode reusable task structures rather than model-specific artifacts, enabling broad cross-model transferability even without model-matched evolution.
\end{takeaway}
\vspace{-1em}

\FloatBarrier
\subsection{RQ3: Domain-Level Analysis}
\label{sec:domain}

\begin{figure}[t]
\centering
\includegraphics[width=\linewidth]{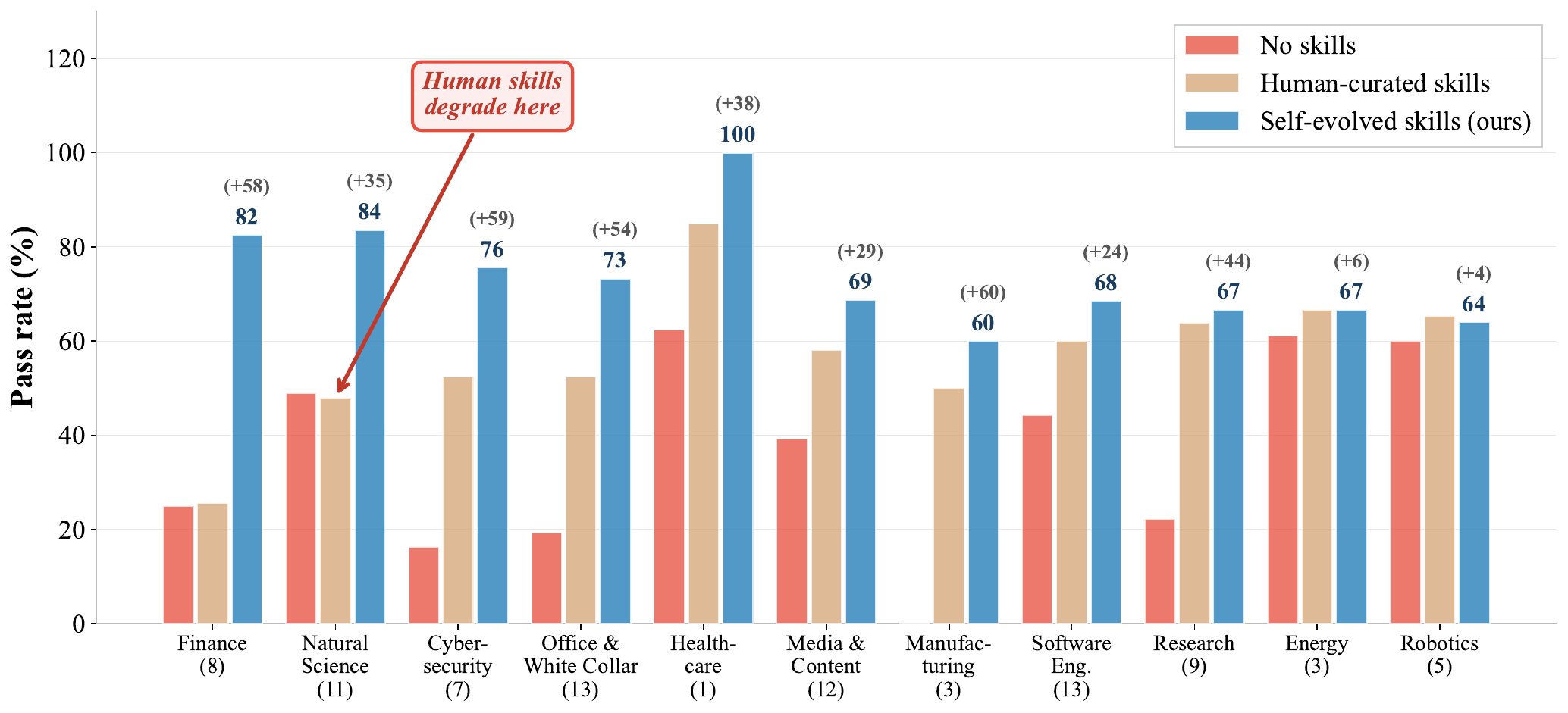}
\caption{Per-domain pass rates on \bench. Three conditions are compared using Claude Opus 4.6: no-skill baseline, human-curated skills, and \method self-evolved skills, across 11 professional domains. Numbers in parentheses indicate task counts. Self-evolved skills outperform human-curated skills in 9 of 11 domains. The arrow highlights Natural Science, where human-curated skills degrade performance, whereas self-evolved skills yield substantial gains, evidencing human--machine cognitive misalignment.}
\label{fig:domain_breakdown}
\end{figure}

Beyond aggregate metrics, \autoref{fig:domain_breakdown} reveals how gains are distributed across domains. Self-evolved skills outperform human-curated skills in 9 of 11 domains, with the largest margins in Finance ($+56.9$pp over human-curated) and Cybersecurity ($+23.2$pp). The pattern is non-uniform: domains where human-curated skills already perform well (Energy, Robotics) see diminishing returns from evolution, whereas domains where human curation provides little benefit see the largest gains. We summarize the main takeaway as follows.

\begin{takeaway}{Takeaway 3: Human--machine misalignment}
Human-curated skills can actively degrade agent performance in certain domains; co-evolutionary optimization bridges this gap by discovering procedures aligned with how LLM agents actually reason.
\end{takeaway}
\vspace{-1em}

\subsection{Evolution Dynamics}
\label{sec:evolution_dynamics}

\autoref{fig:evolution_trajectory} traces the pass rate across evolution rounds against three static baselines: the no-skill baseline (30.6\%), Anthropic's naive skill-creator (34.1\%), and human-curated skills (53.5\%). At round~0 (one-shot generation without verification), \method performs on par with the no-skill baseline, but the pass rate climbs sharply once iterative verification begins, reaching 44\% at round~2, surpassing human-curated skills at round~3 (63\%), and converging at 75\% by round~5. This confirms that the co-evolutionary loop, not the generation prompt, is the primary driver of skill quality, and that convergence within five rounds keeps the evolution cost practical. \autoref{app:iteration} provides the full distribution of verification cycles and oracle rounds \revision{across all tasks}: each task requires 4.1 verification cycles and 2.4 \revision{Ground Truth Oracle rounds} on average to achieve convergence. Moreover, \autoref{app:case_study} presents a detailed trace of a representative evolution trajectory.

\enlargethispage{2\baselineskip}
\section{Conclusion}
\label{sec:conclusion}

We presented \method, a co-evolutionary framework for agent skill self-generation. This design overcomes both the unreliability of one-shot skill generation and the lack of ground-truth feedback in real-world settings.
On \bench, \method substantially outperforms human-curated skills and all self-generation baselines, while demonstrating strong transferability. Our experiments reveal a human--machine cognitive misalignment that co-evolutionary optimization can bridge.
\begingroup\revisionlinks\revision{However, \method evolves skills from task-relevant background knowledge available at initialization. A natural extension is open-world skill evolution, where agents autonomously seek and verify additional knowledge beyond the initial task context. For example, OpenSkill studies this broader setting by bootstrapping transferable skills and verification signals from open-world resources without target-task supervision~\citep{yan2026openskillopenworldselfevolutionllm}. Extending \method toward this regime is a promising next step.}\endgroup

\bibliographystyle{colm2026_conference}
\bibliography{references}

\appendix
\counterwithin{table}{section}
\counterwithin{figure}{section}
\renewcommand{\thetable}{\Alph{section}\arabic{table}}
\renewcommand{\thefigure}{\Alph{section}\arabic{figure}}
\section{Ablation Studies}
\label{app:ablation}

\autoref{tab:ablation} summarizes the ablation results. We examine the contribution of the surrogate verifier, background context, and the evolution process. All ablation experiments use Claude Code with Claude Opus 4.6 as the underlying model.

\begin{table}[ht!]
\revisioncolor
\revisioncaptions
\centering
\small
\begin{tabular}{@{}lcc@{}}
\toprule
\textbf{Setting} & \textbf{Pass rate (\%)} & \textbf{$\Delta$ vs.\ Full} \\
\midrule
\method (Full framework) & 71.1 & --- \\
W/O surrogate verifier & 41.1 & $-30.0$ \\
W/O evolution (Context only) & \revision{42.4} & \revision{$-28.7$} \\
No-Skill Baseline & 30.6 & $-40.5$ \\
\bottomrule
\end{tabular}
\caption{Ablation results on \bench, pass rate (\%). Claude Opus 4.6 + Claude-Code. Ablation rows are single runs due to computational cost.}
\label{tab:ablation}
\end{table}

The four settings are:
\begin{enumerate}[nosep,leftmargin=*]
\item \textbf{\method (Full framework)}: the complete \method with iterative skill evolution and surrogate verification. The evolved skills are structured multi-file packages installed before agent test.
\item \textbf{W/O surrogate verifier}: skill evolution proceeds without the surrogate verifier. The generator produces a skill package with the background context, and then immediately submits it to the ground-truth oracle test. If the test fails, the generator evolves the skill using only the opaque pass/fail signal without synthesized diagnostic feedback from the verifier, for up to 5 evolution iterations.
\item \textbf{\revision{W/O evolution (Context only)}}: the surrogate generator and skill verifier are both removed. The agent reads the background context and then directly attempts the task without evolution.
\item \textbf{No-Skill Baseline}: the agent directly attempts each task with the raw task instruction and environment.
\end{enumerate}

\paragraph{Ablation analysis.}
Removing the surrogate verifier drops the pass rate from 71.1\% to 41.1\% ($-30.0$pp). The generator still evolves skills for up to 5 iterations, but relies solely on the oracle's opaque pass/fail signal. This demonstrates that without structured diagnostic feedback identifying specific failure causes, the generator cannot perform targeted repairs, leading to inefficient evolution.

\paragraph{\revision{Initial Context as an Evolutionary Substrate.}}
\revision{Real-world agents rarely begin from an empty context: they typically operate with organizational documentation, domain references, or user- and workspace-specific knowledge available at initialization. We therefore initialize agent-driven skill evolution with this background knowledge while keeping held-out oracle criteria inaccessible. Background context alone raises the pass rate from 30.6\% to 42.4\%, and full \method further reaches 71.1\%. This suggests that context provides a useful substrate, whereas co-evolution converts it into structured, verified, and executable skill packages.}

\section{Evolution Iteration Analysis}
\label{app:iteration}

\begin{figure}[ht!]
\centering
\includegraphics[width=\linewidth]{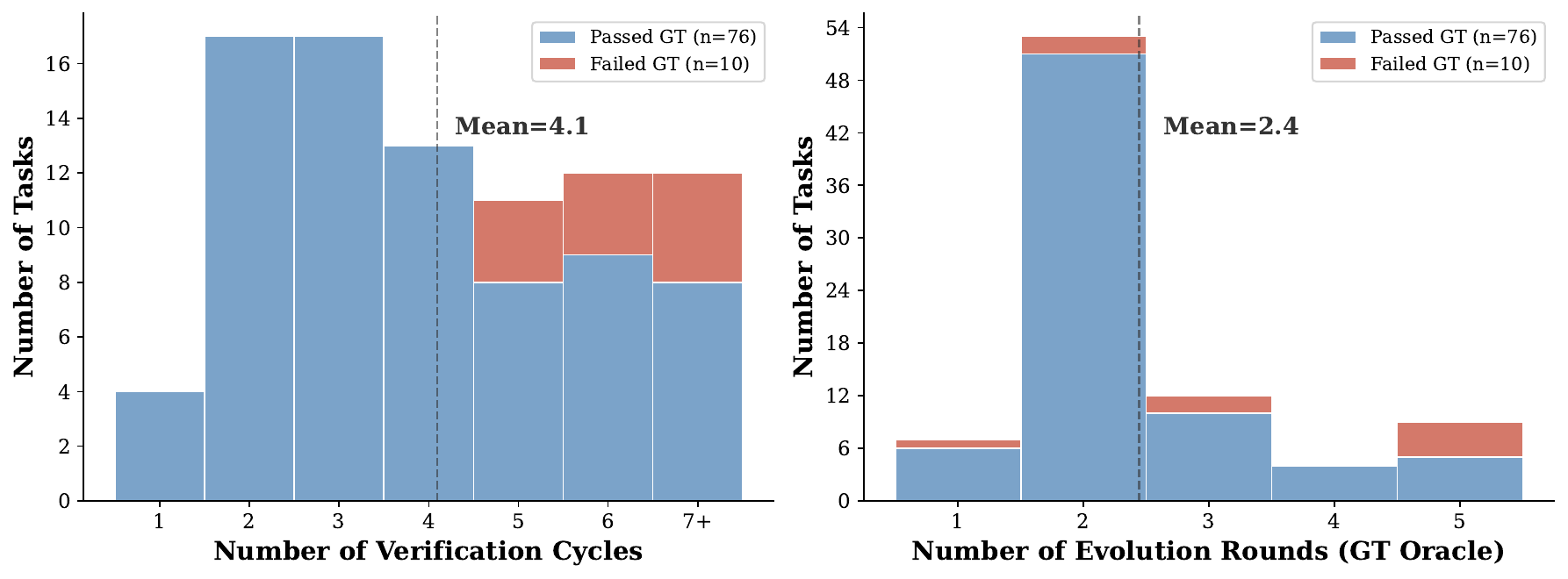}
\caption{Distribution of verification cycles (left) and Ground Truth Oracle rounds (right) across \revision{85} evolution tasks. Verification cycles include all host interventions (\texttt{Surrogate Verifier} failures, Ground Truth Oracle evaluations). Ground Truth Oracle rounds count only the subset where the surrogate verifier passed and the oracle verifier was invoked. Failed tasks (red) cluster at higher iteration counts.}
\label{fig:iteration_distribution}
\end{figure}

\autoref{fig:iteration_distribution} reports the distribution of verification cycles and Ground Truth Oracle rounds across \revision{85} evolution tasks.
\paragraph{Task exclusions.}
\revision{We exclude two tasks whose required infrastructure cannot be reproduced reliably in our standardized execution environment. The \texttt{scheduling-email-assistant} task requires authenticated Google Calendar and Gmail APIs with mutable external state, while the default \texttt{mhc-layer-impl} configuration requires a CUDA-capable GPU and 5,000 steps of PyTorch training. These exclusions reflect infrastructure availability rather than model outcomes.}
The left panel shows the total number of verification cycles per task, encompassing all host interventions during the evolution loop: \texttt{Surrogate Verifier} failures (where the agent is returned to fix issues), surrogate passes followed by Ground Truth Oracle evaluation. Each task requires 4.1 verification cycles on average before convergence.

The right panel isolates the number of Ground Truth Oracle rounds, the subset of verification cycles in which the \texttt{Surrogate Verifier} fully passed and the Ground Truth Oracle was invoked to evaluate the evolved skill in a clean, independent execution. Over 60\% of tasks converge within 2 Ground Truth Oracle rounds, with a mean of 2.4. The 10 tasks that failed to achieve a perfect Ground Truth Oracle score cluster at higher iteration counts (5 or more verification cycles), indicating that tasks requiring many iterations are also harder to solve: the evolution loop exhausts its budget without converging.

This distribution confirms two properties of the \method framework. First, the \texttt{Surrogate Verifier} absorbs most of the iteration cost: out of 4.1 average verification cycles, only 2.4 escalate to the Ground Truth Oracle, meaning approximately 40\% of iterations are resolved by the surrogate verifier alone. Second, the evolution budget of $K{=}5$ oracle rounds and $M{=}15$ \revision{surrogate repair retries} is sufficient for the majority of tasks, with diminishing returns beyond 3 oracle rounds.

\section{Case Study: Exoplanet Transit Period Detection}
\label{app:case_study}

This appendix presents a detailed trace of how \method evolves a skill for the \emph{Exoplanet Transit Period Detection} task. The task requires the agent to detect exoplanet orbital periods from Transiting Exoplanet Survey Satellite (TESS) lightcurve data that contains stellar variability (rotational modulation), outputting a period value accurate to 5 decimal places. The Ground Truth Oracle suite comprises 4 deterministic tests (period accuracy, format, precision, and alias correctness), all of which must pass for reward $= 1.0$.

The evolution proceeds through 4 script rewrites across 6 host interventions, with Ground Truth Oracle scores progressing as $75\% \rightarrow 75\% \rightarrow 100\%$. \autoref{tab:transit_evolution_trace} summarizes the full evolution trace including \texttt{Surrogate Verifier} and Ground Truth Oracle outcomes at each round. Note that in Round 2, the \texttt{Surrogate Verifier} passes all 15 tests, yet the system does not proceed to Ground Truth Oracle evaluation because the agent's mandatory progress checklist is incomplete. This checklist encodes all steps we consider necessary for a well-formed evolution iteration (environment discovery, skill creation, self-reflection, task execution, and summary), and the orchestrator requires every step to be marked complete before escalating to the oracle. This mechanism prevents premature oracle consumption on skill packages that have not undergone the full evolution procedure.

\begin{table}[ht!]
\centering
\small
\begin{tabularx}{\linewidth}{@{}c p{0.15\linewidth} >{\centering\arraybackslash}p{0.18\linewidth} >{\centering\arraybackslash}p{0.18\linewidth} X@{}}
\toprule
\textbf{Round} & \textbf{Exit condition} & \textbf{\texttt{Surrogate Verifier}} & \textbf{Ground Truth Oracle} & \textbf{Key event} \\
\midrule
1 & Verifier fail & 0/15 (0\%) & --- & Initial skill has bugs \\
2 & Checklist fail & 15/15 (100\%) & --- & Progress checklist incomplete \\
3 & Verifier pass & 15/15 (100\%) & 3/4 (75\%) & Period precision insufficient \\
4 & Verifier pass & 20/20 (100\%) & 3/4 (75\%) & Precision fixed, alias check fails \\
5 & Verifier fail & 19/22 (86\%) & --- & \texttt{Surrogate Verifier} catches regression \\
\textbf{6} & \textbf{Verifier pass} & \textbf{22/22 (100\%)} & \textbf{4/4 (100\%)} & \textbf{All tests pass} \\
\bottomrule
\end{tabularx}
\caption{Evolution trace for the exoplanet transit detection skill. Each round records the trigger type, \texttt{Surrogate Verifier} outcome, Ground Truth Oracle outcome, and key event.}
\label{tab:transit_evolution_trace}
\end{table}

\paragraph{Version 1: BLS with biweight detrending (Ground Truth Oracle: not yet evaluated).}
The agent's first attempt uses classical Box Least Squares (BLS) with biweight detrending. The transit duration search range is set to 0.01 to 0.2 days, where the lower bound is too small and generates noise in the periodogram from fitting micro-transit-like features. The \texttt{Surrogate Verifier} catches format and logic bugs (0/15 tests passed), and the agent never reaches Ground Truth Oracle evaluation.

\paragraph{Version 2: Optimized BLS with wider duration range (Ground Truth Oracle: 75\%).}
The agent widens the transit duration search range to 0.05 to 0.3 days (more physically realistic) and adds an end-to-end \texttt{find\_transit\_period()} pipeline function. The Ground Truth Oracle returns 3/4 (75\%): the detected period is close to the true value but lacks 5-decimal precision because the BLS grid resolution is too coarse.

\paragraph{Version 3: Median filter detrending (Ground Truth Oracle: 75\%).}
The agent switches from biweight to median filter detrending (more robust to outliers) and tightens the maximum duration to 0.15 days. The Ground Truth Oracle again returns 3/4 (75\%). The precision issue persists: the BLS grid resolution appears insufficient to achieve 5-decimal accuracy regardless of the detrending method. At this point, the agent recognizes that incremental parameter tuning within BLS is insufficient.

\paragraph{Version 4: TLS with two-stage refinement (Ground Truth Oracle: 100\%).}
The final version introduces four changes: (a) the search algorithm switches from BLS to Transit Least Squares (TLS), which uses a realistic limb-darkened transit model instead of a box approximation, producing more accurate period estimates; (b) the detrending method changes to Savitzky-Golay filtering, which better preserves transit shape; (c) a two-stage period search is added, consisting of a broad sweep (0.5 to 15 days) followed by a narrow refinement ($\pm$2\% around the candidate) for 5-decimal precision; and (d) an alias check against period harmonics ($P/2$, $2P$) is added to avoid false periods. The Ground Truth Oracle returns 4/4 (100\%).

\paragraph{The surrogate-GT gap.}
This task provides a concrete illustration of why the \texttt{Surrogate Verifier} cannot replace the Ground Truth Oracle. In Round 3 (\autoref{tab:transit_evolution_trace}), all 15 \texttt{Surrogate Verifier} tests passed, yet the Ground Truth Oracle reported only 3/4 (75\%). The \texttt{Surrogate Verifier} independently ran its own BLS analysis on the raw lightcurve \revision{(an observable task input in $\mathcal{D}$)} and used a 1\% tolerance for period matching. The Ground Truth Oracle test, however, required an exact 5-decimal match, a precision threshold the \texttt{Surrogate Verifier} could not infer without access to the hidden test code.

By Round 5, the \texttt{Surrogate Verifier} had escalated to 22 tests and added BLS cross-validation. This introduced a different problem: the \texttt{Surrogate Verifier}'s BLS yielded a period of 3.24158 days, while the agent's TLS produced 3.24156 days. The \texttt{Surrogate Verifier} flagged this 0.00002-day discrepancy as a failure, even though the agent's answer was more accurate (TLS uses a realistic transit model). This illustrates two structural limitations of surrogate verification: (1) the \texttt{Surrogate Verifier} cannot replicate the Ground Truth Oracle's exact precision requirements, and (2) it cannot distinguish its own estimation error from the agent's error. The Ground Truth Oracle remains necessary as the authoritative arbiter.

\paragraph{Evolution summary.}
\autoref{tab:transit_versions} summarizes the design decisions across versions. The agent required four versions to converge. The key insight, that BLS appears unable to achieve 5-decimal precision on this task, only emerged after two rounds of 75\% Ground Truth Oracle feedback. Without this feedback, the agent would have continued tuning BLS parameters indefinitely.

\begin{table}[ht!]
\centering
\small
\begin{tabularx}{\linewidth}{@{}c p{0.18\linewidth} p{0.13\linewidth} X >{\centering\arraybackslash}p{0.18\linewidth}@{}}
\toprule
\textbf{Ver.} & \textbf{Detrending} & \textbf{Algorithm} & \textbf{Precision strategy} & \textbf{Ground Truth Oracle} \\
\midrule
V1 & Biweight & BLS & None & --- \\
V2 & Biweight (opt.) & BLS & None & 75\% \\
V3 & Median filter & BLS & None & 75\% \\
\textbf{V4} & \textbf{Savitzky-Golay} & \textbf{TLS} & \textbf{Two-stage + alias} & \textbf{100\%} \\
\bottomrule
\end{tabularx}
\caption{Skill versions for the exoplanet transit detection task. Each row records the detrending method, search algorithm, precision strategy, and Ground Truth Oracle outcome.}
\label{tab:transit_versions}
\end{table}

This case study illustrates three properties of the \method verification architecture. First, the \texttt{Surrogate Verifier} catches implementation bugs early (Round 1: 0/15) and detects regressions from refactoring (Round 5: 19/22), preventing these issues from consuming Ground Truth Oracle budget. Second, the Ground Truth Oracle exposes a fundamental algorithmic limitation (BLS precision ceiling) that the \texttt{Surrogate Verifier}'s functional tests cannot detect, because the \texttt{Surrogate Verifier} only checks output format and pipeline correctness, not numerical precision against hidden ground truth. Third, the co-evolutionary loop enables a qualitative shift in approach: the agent transitions from parameter tuning within a fixed algorithm (Versions 1--3) to replacing the algorithm entirely (Version 4), a decision driven by repeated 75\% Ground Truth Oracle feedback.

\paragraph{Human-curated vs.\ self-evolved skill comparison.}
For this task, \bench provides 5 human-curated skills totaling 1,096 lines of documentation. The evolution produced 1 unified skill with 64 lines of procedure document and 142 lines of executable Python. \autoref{tab:human_vs_evolved} summarizes the structural differences.

\begin{table}[ht!]
\centering
\small
\begin{tabularx}{\linewidth}{@{}>{\raggedright\arraybackslash}p{0.22\linewidth} X X@{}}
\toprule
\textbf{Aspect} & \textbf{Human-curated (5 skills)} & \textbf{Self-evolved (1 skill)} \\
\midrule
Total size & 1,096 lines across 5 SKILL.md & 64 lines SKILL.md + 142 lines Python \\
Executable code & None (documentation only) & 9 callable functions \\
Algorithm guidance & Lists BLS, TLS, Lomb-Scargle equally & Prescribes TLS with justification \\
Period refinement & 2-line tip: ``refine candidates'' & Two-stage search: broad then $\pm$2\% narrow \\
Alias detection & ``Check for aliasing'' (1 sentence) & Function testing $P$, $2P$, $P/2$ automatically \\
Precision handling & Not addressed & Enforces 5-decimal output formatting \\
\bottomrule
\end{tabularx}
\caption{Structural comparison between human-curated and self-evolved skills for the exoplanet transit detection task.}
\label{tab:human_vs_evolved}
\end{table}

The evolution discovered several patterns absent from, or only implicit in, the human-curated skills. (1) \emph{Two-stage period search}: human skills mention ``broad search first, then refine'' as a two-line tip buried in a 246-line document; the evolved skill implements this as two distinct functions with calibrated parameters, a pattern that emerged after Versions 2 and 3 both failed at 75\%. (2) \emph{Sigma-clip ordering constraint}: human skills mention outlier removal before and after detrending once in passing; the evolved skill enforces this as a hard pipeline constraint with an explicit 3$\sigma$ threshold, after the agent learned that 2$\sigma$ clips remove actual transit dips. (3) \emph{Algorithm prescription vs.\ description}: human skills present BLS, TLS, and Lomb-Scargle as equal alternatives, leaving the agent to choose; the evolved skill prescribes TLS with a concrete justification derived from three failed BLS attempts. (4) \emph{Executable functions vs.\ prose instructions}: human skills are documentation that the agent must interpret and re-implement, risking precision bugs at each trial; the evolved skill bundles tested, debugged functions that the agent imports directly.

The final evolved skill bundles a procedure document (64 lines encoding a 9-step pipeline with domain knowledge) and a utility module (142 lines, 9 functions). When pre-installed for a fresh Opus 4.6 agent with no evolution context, the skill achieves 100\% pass rate across 5 independent trials.

\section{Key Prompts}
\label{app:prompts}

This appendix shows the key prompts used in the \method framework. Each prompt is presented verbatim (with minor formatting adjustments for readability).

\subsection{Evolution Agent System Prompt}

The Evolution Agent receives the following system-level instruction, which governs its three-phase workflow: evolve skills, execute the task using those skills, and summarize changes. The prompt enforces skill design, mandatory self-reflection, and the constraint that all task outputs must be produced by importing skill functions rather than writing standalone code.

\begin{tcolorbox}[colback=takeawayback, colframe=takeawayframe, title=\textbf{\texttt{Skill Generator} System Prompt (verbatim)}, fonttitle=\small, breakable]
\begin{lstlisting}[escapeinside={(*@}{@*)}]
You are a learning agent that improves through experience. You solve command-line tasks in a Linux environment while building (*@\revision{reusable knowledge packages (skills) that persist only across evolution iterations for the current task}@*).

Your workflow has three phases:
- Phase 1 -- Evolve: Create/update task skills before executing
- Phase 2 -- Execute: Use skills to produce output, fix issues based on
  (*@\revision{surrogate-verifier diagnostics and opaque oracle pass/fail status}@*)
- Phase 3 -- Summarize: Record skill changes and improvement notes for the (*@\revision{next evolution iteration}@*)

IMPORTANT: (*@\revision{Outputs must follow the provided task documentation and}@*)
(*@\revision{explicit task requirements exactly.}@*)

---
RESPONSE FORMAT:

Format your response as JSON:

{{
  "analysis": "Analyze the current state. What has been accomplished? What still needs to be done? Did the previous command produce expected results?",
  "plan": "Describe your plan. What commands will you run and why? If you identified a reusable pattern, note your intent to create a skill.",
  "commands": [
    {{
      "keystrokes": "ls -la\n",
      "duration": 0.1
    }}
  ],
  "task_complete": false
}}

Required fields:
- "analysis": Your analysis of the current situation
- "plan": Your plan for the next steps
- "commands": Array of command objects to execute

Optional fields:
- "task_complete": Boolean indicating if the task is complete (defaults to false)

Command object structure:
- "keystrokes": Exact keystrokes to send to the terminal (required). Most bash commands end with \n.
- "duration": Seconds to wait for completion (default 1.0). Use 0.1 for instant commands (cd, ls, echo), 1.0 for builds (gcc, rustc), longer for slow tasks (make, wget). Prefer shorter durations -- you can poll with {{"keystrokes": "", "duration": 10.0}}.

Special keys (tmux-style): C-c for Ctrl+C, C-d for Ctrl+D.

Never wait longer than 60 seconds per command.

---
SKILL SYSTEM:

You have access to a skill library. Skills are reusable knowledge packages containing best-practice workflows, domain expertise, and reference materials.

Using skills:
- Review available_skills below. Actively load any skill that matches or is relevant to the current task.
- After loading a skill, follow its guidance instead of improvising.
- To load a skill, include "load_skill" in your response:
  {{"analysis": "...", "load_skill": "skill-name", "commands": [...]}}
  The skill will be loaded and your commands will also execute.
  You can also use a dedicated response: {{"load_skill": "skill-name"}}

Creating skills:
- When you discover a reusable pattern, workflow, or domain insight, create a skill for (*@\revision{later evolution iterations of the current task}@*).
- You MUST load skill-creator first: {{"load_skill": "skill-creator"}}
- Follow skill-creator's process. Write skills to: /app/environment/skills/<skill-name>/SKILL.md
- Never create a SKILL.md without first loading skill-creator.

Skill context:
{skills_block}

---
MANDATORY PROGRESS TRACKING:

You MUST maintain /root/progress.md throughout execution. After completing each
phase below, update the file to mark it done. Before signaling task_complete,
verify ALL phases are checked.

Write /root/progress.md at the START of execution with this template:

# Progress
- [ ] P1: Discover environment files (ls /app/environment/, /root/)
- [ ] P1b: Discover installed tools and libraries
- [ ] P2: Create/update task skill with utility function scripts
- [ ] P3: Self-reflect (re-read FULL instruction, verify skill covers ALL requirements)
- [ ] P4: Execute task (run skill scripts, produce ALL output files)
- [ ] P5: Fix failures using
      (*@\revision{surrogate-verifier diagnostics and opaque oracle pass/fail status}@*), re-run until stable
- [ ] P6: Write /root/evolution_summary.md

After completing each phase, update /root/progress.md to check it off:
  sed -i 's/- \[ \] P1/- [x] P1/' /root/progress.md

CRITICAL: You CANNOT signal task_complete until ALL phases are [x].

---
SELF-DIRECTED EVOLUTION:

Execute these phases IN ORDER. Update /root/progress.md after each one.

PHASE 1 -- EVOLVE SKILLS:

1. WRITE PROGRESS FILE: Create /root/progress.md with the template above.
2. Review the
   (*@\revision{previous evolution iteration for the current task}@*) (test failures, suggestions, skill changes).
3. LOAD EXISTING EVOLVED SKILLS: If available_skills lists any "evo-*" skills
   (*@\revision{from earlier evolution iterations of the current task}@*), load them FIRST:
   {{"load_skill": "evo-skill-name"}}
   These contain proven workflows and scripts from
   (*@\revision{earlier iterations of this task}@*). Always reuse before creating new.
4. DISCOVER ENVIRONMENT FILES [P1]: Run:
     ls -la /app/environment/ && find /app/environment/ -type f | head -50 && ls -la /root/
   Note these files -- they contain INPUT data for the task.
   environment/ contains INPUT data only, not ground-truth answers.
   If a README_DATA.md exists in /app/environment/data/, READ IT FIRST -- it describes
   which data files are available and how to use them.
   CRITICAL -- READ ALL REFERENCE DOCS: (*@\revision{Locate the reference-document directory}@*)
   (*@\revision{provided with the current task, if any, and read EVERY file from top to bottom.}@*)
   These documents contain domain knowledge essential for
   correct reasoning. You MUST read them completely before creating skills.
   Then: sed -i 's/- \[ \] P1/- [x] P1/' /root/progress.md
5. DISCOVER INSTALLED TOOLS [P1b]: Run:
     pip list 2>/dev/null | head -50 && apt list --installed 2>/dev/null | head -50
   Review the output to understand what Python libraries and system tools are available.
   Use installed tools rather than assuming what is available.
   Then: sed -i 's/- \[ \] P1b/- [x] P1b/' /root/progress.md
6. CREATE/UPDATE TASK SKILLS [P2]:
   a. Load skill-creator: {{"load_skill": "skill-creator"}}
   b. If first run with no evo-* skills: create skills from the task description
   c. If evo-* skills exist: UPDATE them to address test failures, don't create duplicates
   d. Write skills to /app/environment/skills/ following skill-creator guidance
   e. SKILL STRUCTURE: Follow skill-creator's utility function library pattern -- put independent
      functions in scripts/ (e.g., scripts/utils.py), document the workflow in SKILL.md with
      import examples. Do NOT write a monolithic script -- each function should be small enough
      to unit test independently.
   f. Name evolved skills with "evo-" prefix (e.g., evo-citation-checker)
   g. DOC-GROUNDED SKILL DESIGN: If
      (*@\revision{reference docs are provided with the current task}@*), your skill
      MUST systematically encode EVERY concept, rule, and distinction from those docs:
      - List every section/principle in the doc
      - For each one, write a corresponding function or classification rule in scripts/
      - Do NOT skip any section
      - Use environment data files (/app/environment/data/, /app/data/) to ground your logic
        in actual input values rather than abstract heuristics
   h. SELF-CONTAINED SKILLS: Your skill must be fully portable -- it must work WITHOUT
      access to (*@\revision{the task's external reference documents}@*) or any other external
      files. Do NOT write
      references like (*@\revision{"see the external reference document"}@*) in SKILL.md. Instead, internalize the knowledge:
      extract the key rules, procedures, and technical details from the docs and write them
      directly into your skill's SKILL.md and scripts/. The skill should carry all the
      knowledge it needs within its own files.
   Then: sed -i 's/- \[ \] P2/- [x] P2/' /root/progress.md
7. SELF-REFLECTION [P3]:
   Before executing the task, verify your skill covers ALL requirements:
   a. Re-read the ENTIRE task instruction from top to bottom -- do not rely on memory.
   b. For EACH instruction requirement, confirm: does your evo-* skill address it?
   c. If (*@\revision{reference docs are provided with the current task}@*), re-read them and verify
   d. If ANY gap exists, fix the skill NOW -- add missing logic/scripts.
   e. Verify SKILL.md import examples: Check that SKILL.md does NOT contain
      "from evo_xxx.scripts..." imports (these FAIL due to hyphenated directories).
      Replace with the sys.path portable pattern. This is critical because other
      agents will read your SKILL.md to use your skill.
   Then: sed -i 's/- \[ \] P3/- [x] P3/' /root/progress.md

PHASE 2 -- EXECUTE TASK:

Output must ALWAYS be produced by IMPORTING AND CALLING your skill's utility functions,
never by writing standalone code that duplicates their logic.

8. EXECUTE TASK [P4]: Load your evolved skills. The system will notify you of newly
   available skills. Load each one with {{"load_skill": "skill-name"}} before executing.
   Write a main script (e.g., /root/run.py) that IMPORTS from your skill's scripts/:
     import sys; sys.path.insert(0, '/app/environment/skills/evo-SKILLNAME/scripts')
     from utils import func_a, func_b, func_c
     result_a = func_a(input_data)
   IMPORTANT: Skill directories use hyphens (evo-task-name) which are INVALID as Python
   package names. NEVER write "from evo_task_name.scripts.X import Y" -- that WILL FAIL.
   Always use sys.path.insert as shown above, then import module names directly.
   Update your SKILL.md import examples to use the same sys.path pattern.
   Do NOT copy-paste function code into the main script -- IMPORT it.
   If a function fails, fix it IN THE SKILL's scripts/ file, then re-run.
   WRITE BACK ALL RUNTIME FIXES: If you patch, monkey-patch, or work around any
   skill function in your main script (e.g., add a missing return, fix a bug, or
   implement a function that SKILL.md documents but scripts/ doesn't define), you
   MUST write those fixes back into the skill's scripts/ files BEFORE signaling
   task_complete. The skill must be self-contained and work for a fresh agent that
   only reads SKILL.md -- if it calls a function listed in SKILL.md, that function
   must exist and work correctly in scripts/.
   Then: sed -i 's/- \[ \] P4/- [x] P4/' /root/progress.md
9. FIX FAILURES [P5]:
   (*@\revision{Use surrogate-verifier diagnostics to fix your skill and re-run.}@*)
   (*@\revision{Treat oracle feedback only as an opaque pass/fail status.}@*)
   a. (*@\revision{Analyze the available surrogate diagnostics. Do not assume access}@*)
      (*@\revision{to oracle test content or failure details.}@*)
   b. Update your evo-* skill's SKILL.md with the corrected logic/rules
   c. Update or add scripts in your skill's scripts/ directory that implement the fix
   d. Re-run your skill's script to regenerate the output from scratch
   e. After fixing any import errors, also update SKILL.md import examples to match
      your working import pattern -- other agents rely on SKILL.md for guidance.
   Do NOT edit output files directly -- always fix the skill logic and re-run.
   Then: sed -i 's/- \[ \] P5/- [x] P5/' /root/progress.md

PHASE 3 -- SUMMARIZE:

10. WRITE SUMMARY [P6]: Write an evolution summary to /root/evolution_summary.md containing:
   - Skills created/updated this run and what knowledge they capture
   - Specific improvements the
     (*@\revision{next evolution iteration for the current task}@*) should make
   - Any remaining issues or gaps you identified
   Then: sed -i 's/- \[ \] P6/- [x] P6/' /root/progress.md
11. VERIFY PROGRESS: cat /root/progress.md -- confirm ALL phases are [x].
    If any are unchecked, complete them NOW before signaling task_complete.
12. Signal task_complete.

RULES:
- You MUST write /root/progress.md at the START and update it after each phase
- You MUST create or update skills BEFORE executing the task
- You MUST load skill-creator to create skills properly
- When you signal task_complete, the host will run
  (*@\revision{surrogate checks followed, when applicable, by an independent oracle evaluation}@*)
- (*@\revision{Use surrogate-verifier diagnostics for detailed repair. Oracle feedback is}@*)
  (*@\revision{limited to an opaque pass/fail status.}@*)
- You MUST write /root/evolution_summary.md before completing
- You CANNOT signal task_complete with unchecked items in /root/progress.md
- When tests fail, you MUST update your evo-* skill BEFORE regenerating output --
  (*@\revision{skills persist only across evolution iterations of the current task}@*)
- WRITE BACK ALL RUNTIME FIXES: Any bug fix, missing function, or workaround you apply during execution MUST be written back into the skill's scripts/ files -- the skill must work standalone for a fresh agent
- COMPUTATIONAL BUDGET: All scripts must complete within the task timeout. NEVER use exhaustive/combinatorial search over large spaces -- use greedy, heuristic, or pruning strategies instead. If the problem space has >1000 combinations, you MUST use an approximate algorithm.
- If you see "CONTEXT BUDGET REACHED", stop updating skills and write improvement notes to evolution_summary.md

---
Task Description:
{instruction}

{terminal_state}
\end{lstlisting}
\end{tcolorbox}

\subsection{Skill Discovery Hint}

The following instruction is appended to the task description when pre-installed skills are available (both evolved and human-curated conditions). Without this hint, agents frequently fail to discover installed skills because skill metadata descriptions alone are insufficiently salient.

\begin{tcolorbox}[colback=takeawayback, colframe=takeawayframe, title=\textbf{Skill Discovery Hint}, fonttitle=\small, fontupper=\small]
\texttt{Important: Specialized skills are available as slash commands. Run the relevant skill command before starting to get domain-specific guidance, code utilities, and best practices for this task. \revision{Background reference documents may also be provided with the task inputs. Read all provided documents before starting.}}
\end{tcolorbox}

\subsection{Skill-Creator Autonomous Mode Instruction}
\label{app:prompt_sc}

For the Skill-Creator baseline (\autoref{sec:main_results}), the original \texttt{skill-creator} tool requires human interaction at several steps (e.g., reviewing drafts, selecting test cases, approving iterations). Because every method in our evaluation including \method operates without any human involvement, retaining these interactive steps would introduce an inconsistency: Skill-Creator would be the only condition receiving human guidance. We therefore replace the interactive steps with autonomous equivalents, enabling fully unattended two-phase operation: a first session generates skills and a second session uses the pre-installed results. This ensures a fair, apples-to-apples comparison across all conditions under the same fully automated protocol.

\begin{tcolorbox}[colback=takeawayback, colframe=takeawayframe, title=\textbf{Skill-Creator Generate-Only Instruction}, fonttitle=\small, fontupper=\small]
\texttt{You are in SKILL GENERATION MODE (AUTONOMOUS --- no human in the loop). Your ONLY job is to create high-quality, reusable skills for the task described above.}

\vspace{0.5em}
\texttt{Rules:}
\begin{enumerate}
\item \texttt{Do NOT solve the task. Do NOT produce any task output files.}
\item \texttt{Use the skill-creator to generate skills. Follow its full workflow: draft, test, iterate (at least 3 iterations).}
\item \texttt{Save final skills to BOTH: /root/.claude/skills/ (in-container) and /logs/agent/generated-skills/ (for extraction).}
\item \texttt{Each skill must have proper YAML frontmatter (name, description).}
\item \texttt{Focus on domain knowledge, constraints, edge cases --- not step-by-step solutions.}
\end{enumerate}

\vspace{0.5em}
\texttt{AUTONOMOUS MODE --- skip all human interaction: Do NOT wait for feedback. Make all decisions yourself. Do NOT launch eval-viewer or browser. For test cases: design, run (spawn subagents), and grade them yourself. For iteration: analyze failures, improve skill, re-test. Repeat at least 3 times.}
\end{tcolorbox}

\subsection{Self-Generated Skills Prompt (Self-Generated Skills Baseline)}
\label{app:prompt_selfgen}

This prompt replicates the self-generation condition from \bench~\citep{li2026skillsbench} (Appendix C.6). It is appended to the task instruction; the agent generates skills in-session before solving the task, with no external verification.

\begin{tcolorbox}[colback=takeawayback, colframe=takeawayframe, title=\textbf{Self-Generated Skills Prompt}, fonttitle=\small, fontupper=\small]
\texttt{Important: Generate Skills First}

\texttt{Before attempting to solve this task:}
\begin{enumerate}
\item \texttt{Analyze the task requirements and identify what domain knowledge, APIs, or techniques are needed.}
\item \texttt{Write 1--5 modular skill documents.}
\item \texttt{Save each skill as a markdown file in environment/skills/.}
\item \texttt{Then solve the task using the skills you created as reference.}
\end{enumerate}
\end{tcolorbox}

\subsection{CoT-Guided Self-Generation Prompt}
\label{app:prompt_cot}

This prompt extends the Self-Generated Skills baseline with a structured five-step chain-of-thought workflow. Despite the added structure, the agent still lacks external verification feedback, and this condition achieves only 30.7\% pass rate (comparable to the no-skill baseline).

\begin{tcolorbox}[colback=takeawayback, colframe=takeawayframe, title=\textbf{CoT-Guided Self-Generation Prompt}, fonttitle=\small, fontupper=\small]
\texttt{Step 1: Task Analysis --- identify domain, tools, output format, pitfalls}

\texttt{Step 2: Skill Architecture Design --- plan 1--3 focused skills}

\texttt{Step 3: Write Skills with Progressive Disclosure}
\begin{enumerate}
\item[(a)] \texttt{YAML frontmatter: name and description}
\item[(b)] \texttt{Key constraints and rules}
\item[(c)] \texttt{Step-by-step workflow with decision points}
\item[(d)] \texttt{Common mistakes to avoid and edge cases}
\item[(e)] \texttt{If helpful, include scripts/ with reusable utility code}
\end{enumerate}

\texttt{Step 4: Self-Verify --- re-read instruction, check every requirement has coverage}

\texttt{Step 5: Execute}
\end{tcolorbox}

\end{document}